\documentclass[runningheads]{llncs}
\usepackage{graphicx}
\usepackage{microtype}
\usepackage{multirow}
\usepackage{booktabs} 
\usepackage{caption}
\usepackage{subfigure}
\usepackage{color, colortbl}
\usepackage{soul}
\usepackage{adjustbox}
\usepackage{tabularx}
\usepackage{orcidlink}
\usepackage{soul}
\begin{document}
\title{Multilingual Detection of Check-Worthy Claims\\using World Languages and Adapter Fusion\thanks{This paper is accepted at ECIR 2023.}}
\titlerunning{\changemarker{Multilingual Detection of Check-Worthy Claims using WL+AF}}
%
\author{Ipek Baris Schlicht\inst{1,2}\orcidlink{0000-0002-5037-2203} \and
Lucie Flek\inst{3}\orcidlink{0000-0002-5995-8454} \and Paolo Rosso\inst{1}\orcidlink{0000-0002-8922-1242}}
\authorrunning{Schlicht et al.}
%
\institute{PRHLT Research Center, Universitat Politècnica de València, Spain \and
DW Innovation, Germany \and
CAISA Lab, University of Marburg, Germany \\
\email{\{ibarsch@doctor,prosso@dsic\}.upv.es}, \email{lucie.flek@uni-marburg.de}\\
}

\authorrunning{Schlicht et al.}
%

\newif\ifproofread

\newcommand{\changemarker}[1]{%
\ifproofread
\textcolor{red}{#1}%
\else
#1%
\fi
}

\newcommand{\changemarkernew}[1]{%
\ifproofread
\textcolor{blue}{#1}%
\else
#1%
\fi
}
\maketitle              
\begin{abstract}
Check-worthiness detection is the task of identifying claims, worthy to be investigated by fact-checkers.
Resource scarcity for non-world languages and model learning costs remain major challenges for the creation of models supporting multilingual check-worthiness detection.

This paper proposes cross-training adapters on a subset of world languages, combined by adapter fusion, to detect claims emerging globally in multiple languages.
(1) With a vast number of annotators available for world languages and the storage-efficient adapter models, this approach \changemarker{is more cost efficient}. Models can be updated more frequently and thus stay up-to-date.
(2) Adapter fusion provides insights and allows for interpretation regarding the influence of each adapter model on a particular language.

The proposed solution often outperformed the top multilingual approaches in our benchmark tasks.

\end{abstract}

\proofreadfalse
\section{Introduction}

There is an increasing demand for automated tools that support fact-checkers and investigative journalists, especially in the event of breaking or controversial news~\cite{HassanQuest2015,DBLP:conf/ijcai/NakovCHAEBPSM21}.
Identifying and prioritizing claims for fact-checking, aka. check-worthy (CW) claim detection, is the first task of such automated systems~\cite{graves2018understanding}. This task helps guiding fact-checkers to potentially harmful claims for further investigation. \changemarker{CW claims, as shown in Table~\ref{tab:example}, are verifiable, of public interest and may invoke emotional responses}~\cite{nakov2021clef}. 
Most studies in this area focus on monolingual approaches, predominantly \changemarker{using English datasets for learning the model}.
Support for multilingualism has become an essential feature for fact-checkers who investigate non-English resources~\cite{ginsborg2021report}. 

\begin{table*}[!ht]
    \captionsetup{font=small}
    \caption{An example of a check-worthy claim from CT21.}
    \centering
    \scriptsize
    \begin{tabular}{p{\textwidth}}
    \toprule
Those who falsely claim Vaccines are 100\% safe and who laugh at anti-vaxxers will be dependent on those who refuse to tell the official lies to say that wide-spread vaccination *is* critical.
23 die in Norway after receiving COVID-19 vaccine: [URL] via [MENTION] \\
\bottomrule
    \end{tabular}
    \label{tab:example}
\end{table*}

Although \changemarker{transformers} achieved competitive results on several multilingual applications~\cite{DBLP:conf/naacl/DevlinCLT19,conneau2020unsupervised} \changemarker{including CW detection~\cite{nakov2021overview,CheckThat:ECIR2022},} there are still two main challenges in the CW task:
(1) Because of the task complexity, there are a few publicly available datasets in multiple languages. Updating a multilingual model to detect even recently emerged claims requires data annotation and timely retraining. Finding fact-checking experts to annotate samples would be hard, especially in low-resourced languages.
(2) Storing standalone copies of each fine-tuned model for every supported language requires vast storage capacities.
Because of the limited budgets of non-profit organizations and media institutes, it would be infeasible to update the models frequently.

World Languages (WLs) are languages spoken in multiple countries, while non-native speakers can still communicate with each other using the WL as a foreign language. English, Arabic and Spanish are examples of WLs \footnote{\url{https://bit.ly/3eMIZ9q}}.
It would appear that finding expert annotators for collecting samples in a WL is easier than finding expert annotators for low-resourced languages.

As a resource-efficient alternative to fully fine-tuning transformer models, adapters~\cite{DBLP:conf/icml/HoulsbyGJMLGAG19,DBLP:conf/icml/Stickland019} have been recently proposed. 
Adapters are lightweight and modular neural networks, learning tasks with fewer parameters and transferring knowledge across tasks~\cite{DBLP:conf/icml/HoulsbyGJMLGAG19,DBLP:conf/icml/Stickland019,DBLP:conf/eacl/PfeifferKRCG21} as well as languages~\cite{DBLP:conf/emnlp/PfeifferVGR20}.
Fine-tuned adapters require less storage than fully fine-tuned pre-trained models.



In this paper, we propose \changemarker{cross-lingual} training of \changemarker{datasets in WLs} with adapters to mitigate resource scarcity and provide a cost-efficient solution.
We first train Task Adapters (TAs)~\cite{pfeiffer2020adapterhub} for each WL and incorporate an interpretable Adapter Fusion (AF)~\cite{DBLP:conf/eacl/PfeifferKRCG21} to combine WL TAs for an effective knowledge transfer among the heterogeneous sources.

Our contributions for this paper are summarized as follows~\footnote{We share our source code at \url{https://bit.ly/3rH6yXu}.}:
\begin{itemize}
    \item We extensively analyze the WL AF model\changemarker{s} on the multilingual CW claim detection task \changemarker{and evaluate the models on zero-shot learning (i.e., the target languages were unseen during training)}. We show that \changemarker{the models} could perform better than monolingual TAs and fully fine-tuned models. They also outperformed the related best performing methods \changemarker{in some languages}. \changemarker{In addition, zero-shot learning is possible with the WL AF models.}
    \item \changemarker{We construct an evaluation to quantify the performance of the models on claims about global/local topics across the languages.} Our approach \changemarker{for curating the evaluation set} could be reused for the assessment of other multilingual, social tasks.
    \item We present a detailed ablation study for understanding the limitations of AF models and their behavior on the CW task.
\end{itemize}

\section{Related Work}

\subsection{Identifying Check-Worthy Claims}
Early studies applied feature engineering to machine learning models and neural network models to identify CW claims in political debates and speeches.
ClaimBuster~\cite{hassan2015detecting,hassan2017toward} was one of the first approaches using a Support Vector Machine (SVM) classifier trained on lexical, affective feature sets.
Gencheva et al.~\cite{gencheva2017context} combined sentence-level and contextual information from political debates for training neural network architectures.
Lespagnol et al.~\cite{DBLP:conf/sigir/LespagnolMU19} employed information nutritional labels~\cite{fuhr2018information} and word embeddings as features.
Vasileva et al.~\cite{DBLP:conf/ranlp/VasilevaAMBN19} applied multitask learning from different fact-checking organizations to decide whether a statement is CW.
Jaradat et al.~\cite{DBLP:conf/naacl/JaradatGBMN18} used MUSE word embeddings to support the CW detection task in English and Arabic.

\changemarker{CheckThat! (CT)} organized multilingual CW detection tasks since 2020~\cite{barron2020overview,nakov2021overview,CheckThat:ECIR2022}. 
They support more languages every year and an increasing number of multilingual systems have been submitted. 
Schlicht et al.~\cite{DBLP:conf/clef/SchlichtPR21} proposed a model supporting all languages in the dataset.
They used a multilingual sentence transformer~\cite{reimers2019sentence} and then fine-tuned the model jointly on a language identification task.
Similarly, Uyangodage et al.~\cite{DBLP:conf/ranlp/UyangodageRH21} fine-tuned mBERT on a dataset containing balanced samples for each language.
Recently, Du et al.~\cite{du2022nus} \changemarker{fine-tuned} mT5~\cite{xue2021mt5} \changemarkernew{on the CW detection task, jointly with multiple related tasks and languages by inserting prompts.} \changemarker{All of the listed methods were limited to the languages they were trained on.}


\changemarker{Kartal and Kutlu~\cite{kartal2022re} evaluated mBERT for zero-shot learning in English, Arabic and Turkish, and observed the performance drop in cross-training. A more effective method would be required for cross-lingual training.}
\changemarker{Furthermore, none of the approaches tackled the resource efficiency issue. In this paper, we use adapters and train them only on WLs for resource efficiency and evaluate unseen languages, leveraging AF to understand which target language in the dataset benefits from transferring knowledge from WLs}. 

\subsection{Adapters}
Adapters have been successfully applied to pre-trained models for efficient fine-tuning to various applications.
Early studies~\cite{DBLP:conf/icml/HoulsbyGJMLGAG19,DBLP:conf/eacl/PfeifferKRCG21} used adapters for task adaptation in English.
Pfeiffer et al.~\cite{DBLP:conf/eacl/PfeifferKRCG21} propose an AF module for learning to combine TAs as an alternative to multi-task learning. 
Some researchers~\cite{ustun-etal-2020-udapter,DBLP:conf/emnlp/PfeifferVGR20,DBLP:conf/emnlp/PfeifferVGR21} exploit language-specific adapters for cross-lingual transfer.
This paper builds upon the works of~\cite{DBLP:conf/eacl/PfeifferKRCG21,DBLP:conf/emnlp/PfeifferVGR21}.
We exploit a cross-lingual training set to learn TAs and then use AF to combine them \changemarker{effectively} and provide interpretability on which source TAs are efficient on the target language. 

\section{Datasets}
\subsection{Task Datasets}
We look\changemarker{ed} for multilingual datasets in WLs and other languages for the experiments. CT21~\cite{DBLP:conf/clef/ShaarHHAHNKKAMB21} and CT22~\cite{CheckThat:ECIR2022} are the only publicly available datasets that meet this requirement. CT21 includes English, Arabic, Turkish, Spanish, and Bulgarian samples.
It deals mainly with Covid-19 events, except for the Spanish samples, which focus only on politics.
\changemarker{Compared with CT21,} CT22 also includes samples in Dutch.
The English, Arabic, Bulgarian, and Dutch samples in the dataset build on the corpora on Covid-19~\cite{alam2021fighting}.
The researchers collected new samples in Turkish and Spanish. The Spanish samples were augmented with CT21. 

The statistics of the datasets are shown in Table~\ref{tab:data_stats}.
English, Spanish and Arabic are the WLs contained in the dataset\changemarker{s}.
Both datasets are imbalanced, i,e. CW samples are under-represented across the languages.
Although English is considered a high-resource language, there are considerably fewer English samples than samples in other languages.
Since some samples of the datasets could overlap, we conducted our research experiments per dataset.

\begin{table*}[!t]
    \begin{minipage}[t]{.5\textwidth}
    \scriptsize
    \captionsetup{font=small}
    \caption{Statistics of CT21 and CT22.}
    \begin{tabular}{llllll}
    \toprule
    & & \multicolumn{2}{l}{\textbf{CT21}} & \multicolumn{2}{l}{\textbf{CT22}} \\
    \midrule
      & \textbf{Split} & \textbf{Total} & \textbf{$\%$CW} & \textbf{Total} & \textbf{$\%$CW}\\
        \midrule
    \multirow{3}{*}{\textbf{ar}} & \changemarker{Train} & 3444 & 22.15 & 2513 & 38.28 \\
       & \changemarker{Dev} & 661 & 40.09 & 235 & 42.55 \\
       & \changemarker{Test} & 600 & 40.33 & 682 & 35.28 \\
    \midrule
    \multirow{3}{*}{\textbf{es}} & \changemarker{Train} & 2495 & 8.02 & 4990 & 38.14 \\
     & \changemarker{Dev} & 1247 & 8.74 & 2500 & 12.20 \\
     & \changemarker{Test} & 1248 & 9.62 & 4998  & 14.09 \\
     \midrule
   \multirow{3}{*}{\textbf{en}} &  \changemarker{Train} & 822 & 35.28 & 2122 & 21.07 \\
    & \changemarker{Dev} & 140 & 42.86 & 195 & 22.56 \\
    & \changemarker{Test} & 350 & 5.43 & 149 & 26.17 \\
    \midrule
    \textbf{tr} & \changemarker{Test}  & 1013 & 18.07 & 303 & 4.62 \\
       \midrule
    \textbf{bg} & \changemarker{Test}  & 357 & 21.29 & 130 & 43.85 \\
        \midrule
    \textbf{nl} & \changemarker{Test}  & - & - & 666 & 47.45 \\
    \bottomrule
    \end{tabular}
    \label{tab:data_stats}
   \end{minipage} %
   \begin{minipage}[t]{.5\linewidth}
    \captionsetup{font=small}
    \caption{Statistics of Global and Local Topics}
   \scriptsize
     \begin{tabular}{llllll}
     \toprule
     && \multicolumn{2}{l}{\textbf{CT21}} & \multicolumn{2}{l}{\textbf{CT22}} \\
     \midrule
     & \textbf{Split} & \textbf{Total} & \textbf{$\%$CW} & \textbf{Total} & \textbf{$\%$CW}\\
    \midrule
     \multirow{2}{*}{\textbf{ar}} & Global & 269 & 43.12 & 116 & 46.55 \\
    & Local & 40 & 42.5 & - & - \\
    \midrule
     \multirow{2}{*}{\textbf{es}} & Global & 1208 & 9.93 & 148 & 22.97 \\
    & Local & - & - & 917 & 12.43 \\
        \midrule
    \textbf{nl}   & Global & - & - & 103 & 56.31 \\
     & Local & - & - & 15 & 46.67 \\
         \midrule
    \textbf{en} & Global & 349 & 5.44 & 14 & 28.57 \\
        \midrule
    \textbf{tr} & Global & 887 & 8.91 & - & -\\
        \midrule
    \textbf{bg} & Global & 356 & 21.35 & 25 & 32\\
     \bottomrule
      \end{tabular}
          \label{tab:data_stats_local}
   \end{minipage}
\end{table*}


\subsection{Topical Evaluation Dataset \label{topical_eval_dataset}}
Some countries are culturally dissimilar or might have different political agendas, thus CW topics might differ among countries.
For example, while vaccination was a globally CW topic throughout the COVID-19 pandemic, some COVID-19 myths were believed only by a few communities~\cite{singh2021misinformation}.
\changemarker{An ideal multilingual system should perform well on global as well as local topics.}

\changemarker{Global topics are the topics present in all languages in all datasets, while local topics are present in only one language.}
We created dedicated datasets to evaluate the performance of the WL models for identifying \changemarker{CW claims across} global and local topics.
This evaluation dataset contains solely global and local topics and was created as follows.
\changemarker{To learn topics across datasets, we first} translated the datasets into English and encoded them with a sentence transformer~\cite{reimers2019sentence}.
\changemarker{Second, we learned a topic modeling on the \changemarker{training} datasets in the WLs by using BERTopic~\cite{grootendorst2022bertopic} to assign topics to test samples in all languages.}
Test samples with topics present in all WL languages were added as presumably global topics to the evaluation set.
BERTopic labeled topics that are unrelated to the \changemarker{learnt} topics with -1.
\changemarker{To select samples with local topics, we chose the samples labeled with -1 from the evaluation and applied a new topic modeling on them.
Test samples with topics that were not present in the test dataset of any other language (i.e., local topics) were added to the evaluation set for local topics.}
The statistics of the local and global evaluation sets are presented in Table~\ref{tab:data_stats_local}.



\section{Methodology}
\begin{figure}[t]
    \centering
    \includegraphics[width=0.4\textwidth]{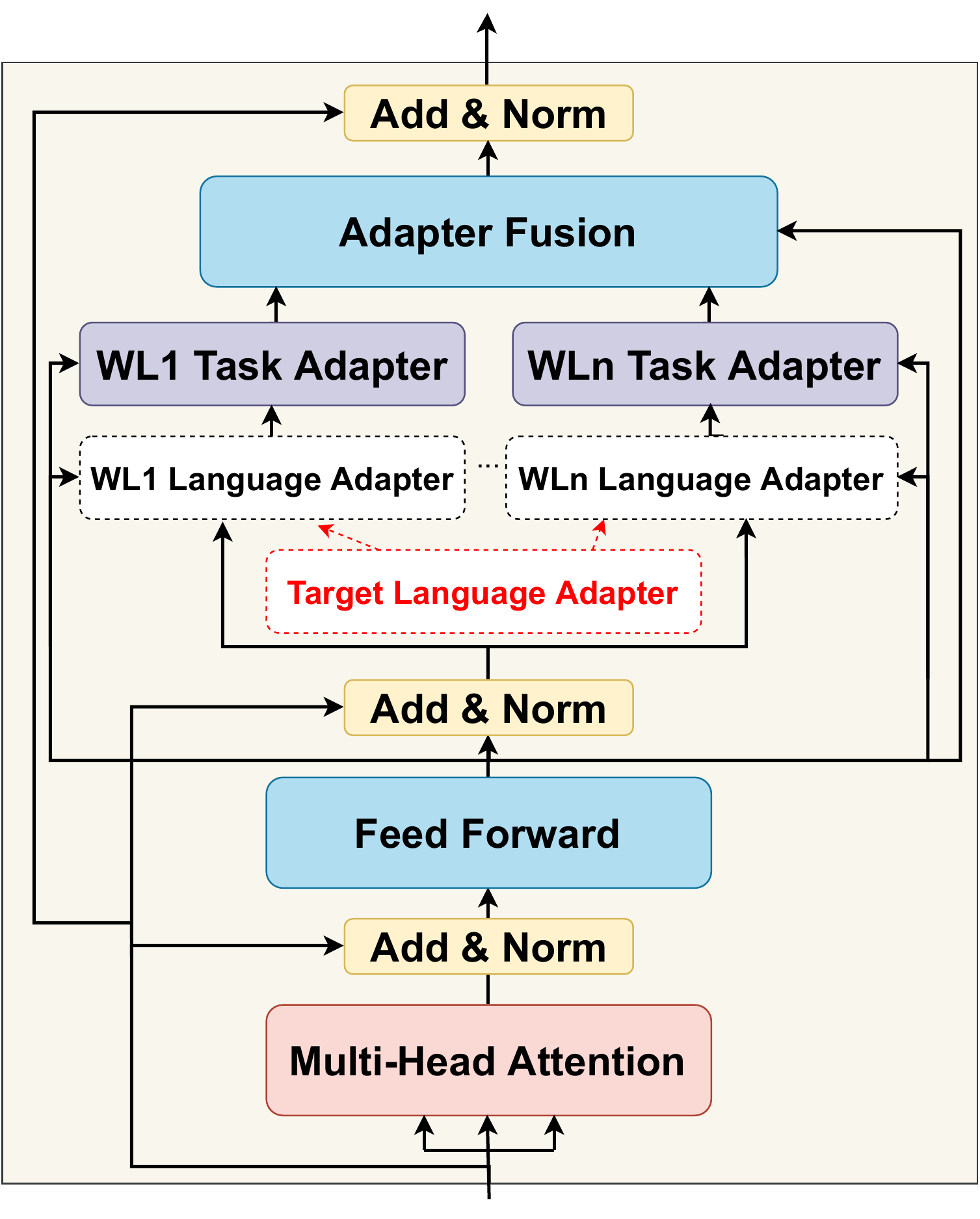}
    \captionsetup{font=small}
    \caption{The architecture of the WL+AF models \changemarker{within a transformer}. WL+AF+LA is the setup when the LAs (blocks with the dashed lines) are stacked into the task adapters. }
    \label{proposed_arch}
\end{figure}

\subsection{World Language Adapter Fusion}
This section describes the WL+AF models.
\changemarker{The} WL+AF models are transformers containing additional modules called adapters.
During training, only the adapters are trained and the weights of pre-trained model are frozen.
Therefore, it is a parameter-efficient approach.
We experiment with two types of WL+AF models.
WL+AF is a standard setup for combining \changemarker{WL} task adapters.
WL+AF+LA has additional language adapters (LAs)~\cite{DBLP:conf/emnlp/PfeifferVGR20} by stacking task adapters.
We illustrate the architecture of the models in Figure~\ref{proposed_arch}.
The input of the architecture is a text while the output is a probability score describing the check-worthiness of the input.


\noindent \\
\textbf{Transformer Encoder.} To provide cross-lingual representations, we use multilingual BERT (mBERT)~\cite{DBLP:conf/naacl/DevlinCLT19} and XLM-Roberta (XLM-R)~\cite{conneau2020unsupervised}\footnote{We use the base version of the model, which consists of 12 layers.}. Because the transformers were previously used by the related studies~\cite{DBLP:conf/ranlp/UyangodageRH21,alam2021fighting}.
Furthermore, there are publicly available LAs for the transformers.
Before encoding text with transformers, we first clean it from URLs, mentions and retweets, then truncate or pad it to 128 tokens.
\noindent \\
\textbf{Task Adapter\label{task_adapter}.}
\changemarker{The annotations might be affected by the different cultural backgrounds and events across the countries, like the CW claims.
Additionally, some may have been created from journalists following manual fact-checking procedures while others stem from crowd-sourcing~\cite{nakov2021overview,CheckThat:ECIR2022}}
For these reasons, we treat each WL dataset as a separate task. 
Then, we obtain TAs by optimizing on their corresponding WL dataset.
TAs consist of one down projection layer followed by a ReLU activation and one up projection layer~\cite{pfeiffer2020adapterhub}.
The TA of each WL is fine-tuned on its corresponding dataset.
\noindent \\
\textbf{Adapter Fusion.}
To share knowledge from the different TAs for predicting samples in an unseen language or new topics, we need to combine them effectively.
AF is a method for combining multiple TAs, and mitigating the common issues of multi-task learning, such as catastrophic forgetting and complete model retraining for supporting new tasks~\cite{DBLP:conf/eacl/PfeifferKRCG21}.
AF consists of an attention module which learns how to combine knowledge from the TAs dynamically.
By analyzing the attention weights of the AF, we could learn which source task adapter contributes more to \changemarker{test predictions}.
We combine the TAs trained on the WLs with an AF.
Then, the AF is fine-tuned on mixed datasets of the WLs to use for the target languages.
\noindent \\
\textbf{Language Adapter.\label{lang_adapter}} To learn CW claim detection from various cross-lingual sources, we should ensure that the model does not learn a language classification task but learns how to identify CW claims.
Therefore, we need to separate language-specific knowledge from task-specific knowledge.
A recent study~\cite{DBLP:conf/emnlp/PfeifferVGR20} demonstrated that LAs achieved better results when transferring the knowledge of a task \changemarker{performed} in one source language into another language.
To encode language-specific information into the transformer model and to see the LAs impact on the performance of the AF models, we use the LAs from Adapter Hub~\cite{pfeiffer2020adapterhub}.
The architecture of the LAs is analog to those of the TAs.
However, these LAs were pre-trained on the unlabeled Wikipedia~\cite{DBLP:conf/emnlp/PfeifferVGR20} by using a masked language model objective.
While learning task-specific transformations with TAs, each world LA is stacked on its corresponding TA.
The weights of the LAs are kept frozen, \changemarker{so they are not learned again}.
During inference of the target task, the source LA is replaced with the target LA.

\subsection{Implementation Details}
We download the pre-trained transformers from Huggingface~\cite{wolf-etal-2020-transformers}.
We fine-tune the TAs on English, Arabic, and Spanish training samples (WLs), and evaluate the models for their capability of zero-shot learning by testing on the other languages.
As LAs for Arabic, English, Spanish and Turkish samples in the dataset~\cite{DBLP:conf/emnlp/PfeifferVGR20}, we download the pre-trained LAs from Adapter Hub~\cite{pfeiffer2020adapterhub} for both transformers.
However, a LA for Bulgarian and Dutch is not available in Adapter Hub, therefore, we use the English LA.

We use the trainers of AdapterHub~\cite{pfeiffer2020adapterhub} for both full fine-tuning and adapter tuning.
We set the number of epochs as 10 and train the models with a learning rate of 1e-4 for CT21 and 2e-5 for CT22\footnote{2e-5 gives better results on the development set of CT22}, and batch size of 16.
We use the best models on the development set according to the dataset metrics.
We repeat the experiments ten times using different random seeds using an NVIDIA GeForce RTX 2080.

\section{Baselines}
We compare the performance of the WL AF models against various baselines:
\begin{itemize}
\item \textbf{Top performing systems:} \changemarker{We chose the top-performing systems on \changemarkernew{CT21 and CT22 which used a single model for the multilingual CW detection instead of containing language-specific models.}
Schlicht et al.~\cite{DBLP:conf/clef/SchlichtPR21} is a runner-up~\footnote{BigIR is the state of art approach, but there is no associated paper/code describing the system.} system on CT21.
The original implementation uses a sentence transformer and the model was fine-tuned on all the language datasets.
For a fair comparison, we set the language identification task to WLs and replaced the sentence transformer with mBERT and XLM-R.
The state-of-the-art model on CT22 is based on mT5-xlarge~\cite{du2022nus}, a multi-task text-to-text transformer.
Like Schlicht et al., the model is fine-tuned on all of the corpora by using multi-task learning.
Due to the limited computing resources, we couldn't fine-tune this model on WLs alone.
We report the results from ~\cite{du2022nus}.}

\item \textbf{Fully fine tuned (FFT) Transformers (on Single Language):} \changemarker{We fine-tune the datasets on a single language of the WLs to evaluate the efficiency of cross-lingual learning: AR+FFT, EN+FFT, and ES+FFT. We also add BG+FFT, ES+FFT, and NL+FFT as the baseline for zero-shot learning.}

\item \textbf{Task Adapters (on Single Language):} \changemarker{These baselines contain a task adapter followed by a LA, a widely used setup for cross-lingual transfer learning with adapters~\cite{DBLP:conf/emnlp/PfeifferVGR20}: AR+TA+LA, EN+TA+LA, ES+TA+LA. Comparing our model to these baselines can help understand whether cross-training with WLs is efficient.}

\item \textbf{Other WL Models:} \changemarker{We evaluate the AF models against a model containing a task adapter trained on WLs (WL+TA). In addition to this baseline, to see if we need a complex fusion method, we use WL+TA+LA+Mean, which takes the average of the predictions by AR+TA, EN+TA and ES+TA. Finally, we analyze the adapter tuning on multiple WLs against the fully fine tuning of mBERT: WL+FFT.}
\end{itemize}

\section{Results and Discussion\label{experiments}}
\begin{table*}[!t]
\captionsetup{font=small}
    \caption{\changemarker{MAP scores for CT21 and F1 scores for CT22 }of the CW detection in WLs. The \textbf{bold} indicates the best score and \underline{underline} indicates the second best score. Overall the AF models performed well on multiple languages while the performance of other models are sensitive to the characteristics of the \changemarker{training set}.}
    \centering
    \scriptsize
    \begin{tabular}{lllllllllllllll}
    \toprule
     & &\multicolumn{3}{l}{\textbf{CT21}} & \multicolumn{3}{l}{\textbf{CT22}} \\
    \midrule
    & & \textbf{ar} & \textbf{es} & \textbf{en} & \textbf{ar} & \textbf{es} & \textbf{en} & \textbf{avg} \\
    \midrule
    & \textbf{Du et al.\cite{du2022nus}} & - & - & - & \textbf{62.8} & 57.1 & \textbf{51.9} & - \\
    \midrule
    \multirow{11}{*}{\textbf{mBERT}}& \textbf{AR+FFT} & 50.17 & 15.78 & 6.03  & \underline{55.52} & 17.85 & 42.92 & 31.38 \\
    & \textbf{ES+FFT} & 41.22 & 20.30 & 6.80 &  37.30 & 54.05 & 45.28 & 34.16 \\
    & \textbf{EN+FFT} & 51.63 & 15.90 & 10.80 &  40.97 & 22.49 & 44.29  & 31.01 \\
    \cmidrule{2-9}
    & \textbf{AR+TA+LA} & 58.20 & 19.97 & 8.62 & 18.13 & 17.61 &  45.73 & 28.04 \\
    & \textbf{ES+TA+LA} & 48.07  & 18.93 & 11.06 & 37.30 & 54.05 &  45.73 & 35.86  \\
    & \textbf{EN+TA+LA} & 50.81 & 40.81 & \textbf{21.21} & 56.39 & 24.63 & 12.93 & 34.46 \\
    \cmidrule{2-9}
    & \textbf{WL+FFT} & 47.93 & 51.50 & 13.85  & 51.50 & 63.20 & 39.94 & \underline{44.65} \\
    & \textbf{Schlicht et al.\cite{DBLP:conf/clef/SchlichtPR21}} & 51.51 & 31.04 & 7.87 & 45.93 & \underline{66.48} & 34.18 & 39.50 \\
    \cmidrule{2-9}
    & \textbf{WL+TA} & 53.77 & 46.58 & 14.44 & 39.54 & 62.69 & 37.19 & 42.37  \\
    & \textbf{WL+TA+LA+Mean} & 54.89 & 35.72 & 12.96 & 0.00 & 33.21 & 51.03 & 31.30 \\
    \cmidrule{2-9}
    & \textbf{WL+AF} & 55.13  & 46.29 & 16.05 & 36.45 & 64.32 & 39.73 & 42.96  \\
    & \textbf{WL+AF+LA} & \underline{55.32} & 46.58 & 15.66 & 39.87 &  \underline{64.64} & 37.27 & 43.22 \\
    \midrule
    \multirow{11}{*}{\textbf{XLM-R}} & \textbf{AR+FFT} & 43.55 & 12.80 & 5.88 & 38.72 &6.83& 41.29 & 24.85 \\
    & \textbf{ES+FFT} & 41.22 & 15.90 & 6.80 &40.25 &21.69 &43.51 & 28.23 \\
    & \textbf{EN+FFT} &43.64 &10.49 & 5.64 &31.46& 64.66 &45.69 & 33.60  \\
    \cmidrule{2-9}
    & \textbf{AR+TA+LA} & 58.16& 25.87 & 7.64 &6.93 &0.06 &1.40 & 16.68 \\
    & \textbf{ES+TA+LA} & 50.27 & \textbf{52.51} & 11.53& 41.24 &65.45 &38.79 & 43.30 \\
    & \textbf{EN+TA+LA} & 56.39 & 24.63 & 12.93& 12.82& 0.06& 28.74 & 22.60\\
     \cmidrule{2-9}
    & \textbf{WL+FFT} &47.93 &13.85& 6.37 &44.53 & 64.75 & \underline{50.93} & 38.06\\
    & \textbf{Schlicht et al.\cite{DBLP:conf/clef/SchlichtPR21}} &51.56 &21.61 &7.32 &42.59 & \textbf{67.13} &36.40 & 37.77 \\
    \cmidrule{2-9}
    & \textbf{WL+TA} &58.02 & \underline{50.76} &11.53 & 42.91 &63.36 &31.69 & 43.05 \\
    & \textbf{WL+TA+LA+Mean} & \textbf{59.32} & 46.11 & 10.49 & 25.32 & 35.55 & 35.28 & 35.35  \\
    \cmidrule{2-9}
    & \textbf{WL+AF} & 58.39 & 49.42 & 13.29 &39.84& 65.66 & 46.96 & \textbf{45.59} \\
    & \textbf{WL+AF+LA} & \underline{58.83} & 47.26 & \underline{16.06} & 35.17 & 65.80 &43.46 & 44.43 \\

    \bottomrule
    \end{tabular}
    \label{tab:results_wl}
\end{table*}

\begin{table*}[!t]
\captionsetup{font=small}
\caption{MAP for \textbf{CT21} and F1 for \textbf{CT22} of the CW detection in zero-shot languages. The \textbf{bold} indicates the best score and \underline{underline} indicates the second best score. The AF models performed well, even outperformed WL+FFT and Schlicht et al.~\cite{DBLP:conf/clef/SchlichtPR21} and some of the monolingual approaches \changemarker{in terms of average score}.  }
    \centering
    \scriptsize
    \begin{tabular}{lllllllllllllll}
    \toprule
     & & \multicolumn{2}{l}{\textbf{CT21}} & \multicolumn{3}{l}{\textbf{CT22}} \\
    & & \textbf{tr} & \textbf{bg} & \textbf{tr} & \textbf{bg} & \textbf{nl} & \textbf{avg} \\
     \midrule
    & \textbf{Du et al.\cite{du2022nus}} & - & - & 17.3 & 61.7 & \textbf{64.2} & - \\
    \midrule
    \multirow{14}{*}{\textbf{mBERT}}& \textbf{BG+FFT} & - & 35.64 & - & 57.16 & - & - \\
    & \textbf{TR+FFT} & 28.47 & - & 14.66 & - & - & -\\
    & \textbf{NL+FFT} & - & - & - & - & \underline{49.80} & -\\
    \midrule
    & \textbf{AR+FFT} & 30.14 & 22.12& 10.71& 54.84 &45.55& 32.67\\
    & \textbf{ES+FFT} & 23.17 & 24.15& 8.19 &47.67 &33.82& 27.4 \\
    & \textbf{EN+FFT} & 42.80 & 33.20 & 13.57 &54.58 &54.32& 39.69 \\
    \cmidrule{2-8}
    & \textbf{AR+TA+LA} & 58.08 & 26.76 &8.04 & 57.43 &58.99 & 41.86 \\
    & \textbf{ES+TA+LA} & 49.09 & 28.09 &8.58 &47.67& 42.16& 35.12\\
    & \textbf{EN+TA+LA} & 54.16 & 44.98 &8.19 &33.92& 33.82 & 35.01\\
    \cmidrule{2-8}
    & \textbf{WL+FFT} & 27.61 & 24.29 & 10.90 & 58.53 &29.03& 30.07 \\
    & \textbf{Schlicht et al.\cite{DBLP:conf/clef/SchlichtPR21}} & 27.81 & 24.41& 7.86 &51.94 &29.33& 28.27\\
    \cmidrule{2-8}
    & \textbf{WL+TA} & 54.11 & 37.04& 9.73&48.02 &36.95 & 37.17\\
    & \textbf{WL+TA+LA+Mean} &  62.32 & 34.52 & 12.02 & 47.17 & 39.91 & 39.19 \\
    \cmidrule{2-8}
    & \textbf{WL+AF} & 50.46 & 39.80 &9.63 &53.55 &38.76& 38.44 \\
    & \textbf{WL+AF+LA} &  50.94 & 40.27 &9.73& 52.75& 43.07& 39.35 \\
    \midrule
    \multirow{11}{*}{\textbf{XLM-R}} & \textbf{BG+FFT} & - & 24.68 & - & 43.47 & - & - \\
    & \textbf{TR+FFT} & 24.57 & - & \textbf{18.90} & - & - & - \\
    & \textbf{NL+FFT} & - & - & - & - & 58.61 & - \\
    \midrule
    & \textbf{AR+FFT} & 23.40 & 21.86 & 11.62 &45.10 &38.73 & 28.14 \\
    & \textbf{ES+FFT} &23.17 &24.15 &9.07 &62.72& 31.24& 30.07  \\
    & \textbf{EN+FFT} &22.63 &21.76 &18.43 &49.25 & 45.62 & 31.54\\
     \cmidrule{2-8}
    & \textbf{AR+TA+LA} &56.19 &21.37 &0.48& 9.75 &5.22& 18.60\\
    & \textbf{ES+TA+LA} &44.98 &23.86& 8.04 &46.72& 30.52 & 30.82 \\
    & \textbf{EN+TA+LA} &58.38& 41.61 &15.19 &8.42 &25.83 & 29.89 \\
    \cmidrule{2-8}
    & \textbf{WL+FFT} & 27.61& 24.29& 14.02 &63.79 &36.34 & 33.21\\
    & \textbf{Schlicht et al.\cite{DBLP:conf/clef/SchlichtPR21}} & 25.86 &22.19& 9.55 &53.75 &24.44 & 27.16 \\
    \cmidrule{2-8}
    & \textbf{WL+TA} & \underline{59.37} &39.72 & 15.60 & \textbf{66.14} &35.98& 43.36 \\
    & \textbf{WL+TA+LA+Mean} & 63.65 & 32.28 & 9.90 & 31.27 & 19.91 & 31.40 \\
    \cmidrule{2-8}
    & \textbf{WL+AF} & 57.46& \underline{46.86} &12.73 &59.12 & 40.83 & \underline{43.4} \\
    & \textbf{WL+AF+LA} & \textbf{61.74} & 41.78 & \underline{17.77} & \underline{63.88} &37.59& \textbf{44.55} \\
    \bottomrule
    \end{tabular}
    \label{tab:results_zl}
\end{table*}

\begin{table*}[!t]
    \captionsetup{font=small}
    \caption{F1 scores of the models on global topics for each dataset. Adapter training is more effective than fully fine-tuning. Although WL+TA outperformed the AF models in particular languages, at average the AF models performed better. }
    \centering
    \scriptsize
    \adjustbox{width=\textwidth}{
    \begin{tabular}{llllllllllll}
    \toprule
    & \multicolumn{5}{l}{\textbf{CT21}} & \multicolumn{6}{l}{\textbf{CT22}} \\
    \midrule
    & \textbf{tr} & \textbf{es} & \textbf{ar}& \textbf{en}   & \textbf{bg}  & \textbf{es} & \textbf{ar} & \textbf{en} & \textbf{bg} & \textbf{nl} & \textbf{avg} \\
    \midrule
    \textbf{WL+FFT} & 0.00 & 0.00 & 2.60 & 0.59 & 0.10 & 77.59 & \textbf{52.13} & \textbf{52.05} & \textbf{48.52} & 38.53 & 27.21 \\ 
    \textbf{Schlicht et al.\cite{DBLP:conf/clef/SchlichtPR21}} & 18.15 & 17.92 & 58.84 & 7.18 & 19.75 &  \textbf{81.81} & 48.12 & 28.33 & 36.55 & 30.34 & 34.70 \\
    \midrule
    \textbf{WL+TA+LA} & \textbf{52.37} & 38.09 & 61.01 & \textbf{13.58} & \textbf{38.44} & 77.28 & 45.47 & 22.57 & 41.31 & \textbf{45.09} & 38.52 \\
    \midrule
    \textbf{WL+AF} &   45.02  & \textbf{41.57} & 61.51 & 13.36 & 36.23 &  79.14 & 44.61 & 50.96 & 40.58 & 43.19 & \textbf{45.62} \\
    \textbf{WL+AF+LA} & 48.52  & 37.44 & \textbf{61.79} & 13.38 & 28.72 &  77.95 & 42.84 & 44.23 & 41.09 & 43.99 & 44.00 \\
    \bottomrule
    \end{tabular}}
    \label{tab:results_global}
\end{table*}

\begin{table*}[!t]
    \captionsetup{font=small}
    \caption{F1 scores of the models on local topics for each dataset. The AF models show similar results to WL+TA and outperformed WL+FFT. }
    \centering
    \scriptsize
    \begin{tabular}{lllll}
    \toprule
    & \textbf{CT21} & \multicolumn{2}{l}{\textbf{CT21}} \\
    \midrule
    & \textbf{ar}& \textbf{es} & \textbf{nl} & \textbf{avg} \\
    \midrule
    \textbf{WL+FFT} & 0.93 &  60.64 & 31.11 & 30.89 \\ 
    \textbf{Schlicht et al.\cite{DBLP:conf/clef/SchlichtPR21}} & 46.80 & \textbf{64.65} & 14.44 & 41.96 \\
    \midrule
    \textbf{WL+TA+LA} &\textbf{50.88} & 61.65 & 38.47 & \textbf{50.33} \\
    \midrule
    \textbf{WL+AF} &  48.15 & 62.01 & 34.38 & 48.18 \\
    \textbf{WL+AF+LA} & 46.41 & 63.07 & \textbf{41.16} & 50.21 \\
    \bottomrule
    \end{tabular}

    \label{tab:results_local}
\end{table*}
\begin{table}[!t] 
    \captionsetup{font=small}
    \caption{Number of training parameters and file size comparisons for the models. \changemarker{mT5 is larger than mBERT and XLM-R.}}
    \scriptsize
    \centering
    \begin{tabular}{llll}
        \toprule
        \textbf{Model} &  \textbf{Base Model} & \textbf{Parameters} & \textbf{Model Size}\\
        \midrule
        \textbf{WL+FFT} & mBERT & 178 M & 711.5 MB \\
                       & XLM-R & 278 M & 1.1 GB \\
        \midrule
        \textbf{Schlicht et al.\cite{DBLP:conf/clef/SchlichtPR21}} & mBERT & 179 M & 716.3 MB \\
                     & XLM-R & 279 M & 1.1 GB \\
        \midrule
        \textbf{TA} \& \textbf{WL+TA+LA} & mBERT & 1.5 M & 6 MB \\
                    & XLM-R & 1.5 M & 6 MB \\
       \textbf{AF} & mBERT & 22 M & 87.4 MB \\
                   & XLM-R & 22 M & 87.4 MB \\
       \textbf{LA} & mBERT & - & 147.78 MB \\
                   & XLM-R & - & 147.78 MB \\
        \midrule & \changemarkernew{mT5} & \changemarkernew{3.7 B}& \changemarkernew{15 GB} \\
        \bottomrule
    \end{tabular}

    \label{tab:size_comparision}
\end{table}


In this section, we present and analyze the results of the WL AF(+LA) models. \changemarker{We compare the models performance at CW detection for (1) WLs, (2) zero-shot languages and (3) local and global topics. Lastly, we compare the performance of WL+AF and WL+AF+LA to investigate whether LA is effective in model performance.}

\changemarker{As seen in Table~\ref{tab:results_wl}, the models trained on single languages are able to perform well for other WLs if only provided with training sets of considerable size, or language of the training and test sets are same. Additionally, Schlicht et al.\cite{DBLP:conf/clef/SchlichtPR21} and WL+FFT were performing well only on CT22, overall, the AF models, \changemarkernew{perform well for various languages}. Du et al.~\cite{du2022nus} outperformed the AF models for Arabic and English samples of CT22, but it underperformed for the Spanish samples. }

As shown in Table~\ref{tab:results_zl}, the AF models achieve good results on target sets in zero-shot languages. \changemarker{It shows that the fusion of multiple sources with adapters could be beneficial in knowledge transfer and is better than the other fusion method WL+TA+LA+Mean. It is noteworthy that Du et al~\cite{du2022nus} was trained on all samples of the training datasets and hence has no zero-shot learning capacity. Although Du et al. achieved a better performance on the Dutch samples, the AF models could obtain similar results in other languages. In terms of resource efficiency, the AF models required less space than WL+FFT and mT5 for storing new weights, as shown in Table~\ref{tab:size_comparision}, which make them more suitable than updating mT5 for newsrooms with a limited budget.}


We compare the performance of models trained on multiple WLs for identifying CW claims about global or local topics. We tested this experiment with the evaluation set described in Section~\ref{topical_eval_dataset} in terms of F1 score. We take the average of the scores of the models coded with mBERT and XLM-R and present them in Tables~\ref{tab:results_global} and \ref{tab:results_local}, respectively, for global and local topics. Overall, the AF models performed better than WL+FFT and Schlicht et al.\cite{DBLP:conf/clef/SchlichtPR21} for both types. However, WL+TA performed similarly to WL+AF+LA in predicting local statements in Arabic samples in CT21. 

Last, we compare the performance of WL+AF and WL+AF+LA to investigate whether LA is effective in model performance. We computed the Fleiss Kappa scores of the AF models for each experiment and language. The overall score is 0.63, which is a moderate agreement. We further investigate the disagreements where the kappa is below 0.5. The conflicts mainly occurred in the zero shot languages and English, with the lowest CW samples on both datasets. Since sometimes WL+AF+LA is better than WL+AF and vice versa, we conclude that LA is not effective in our experiments. The pre-trained LAs were trained on the Wikipedia texts~\cite{DBLP:conf/emnlp/PfeifferVGR20}. Thus, they might miss the properties of social texts, which are mostly noisy.

\begin{figure*}[!t]
    \scriptsize
    \centering
    \subfigure[CT21]{
    \includegraphics[width=0.25\textwidth]{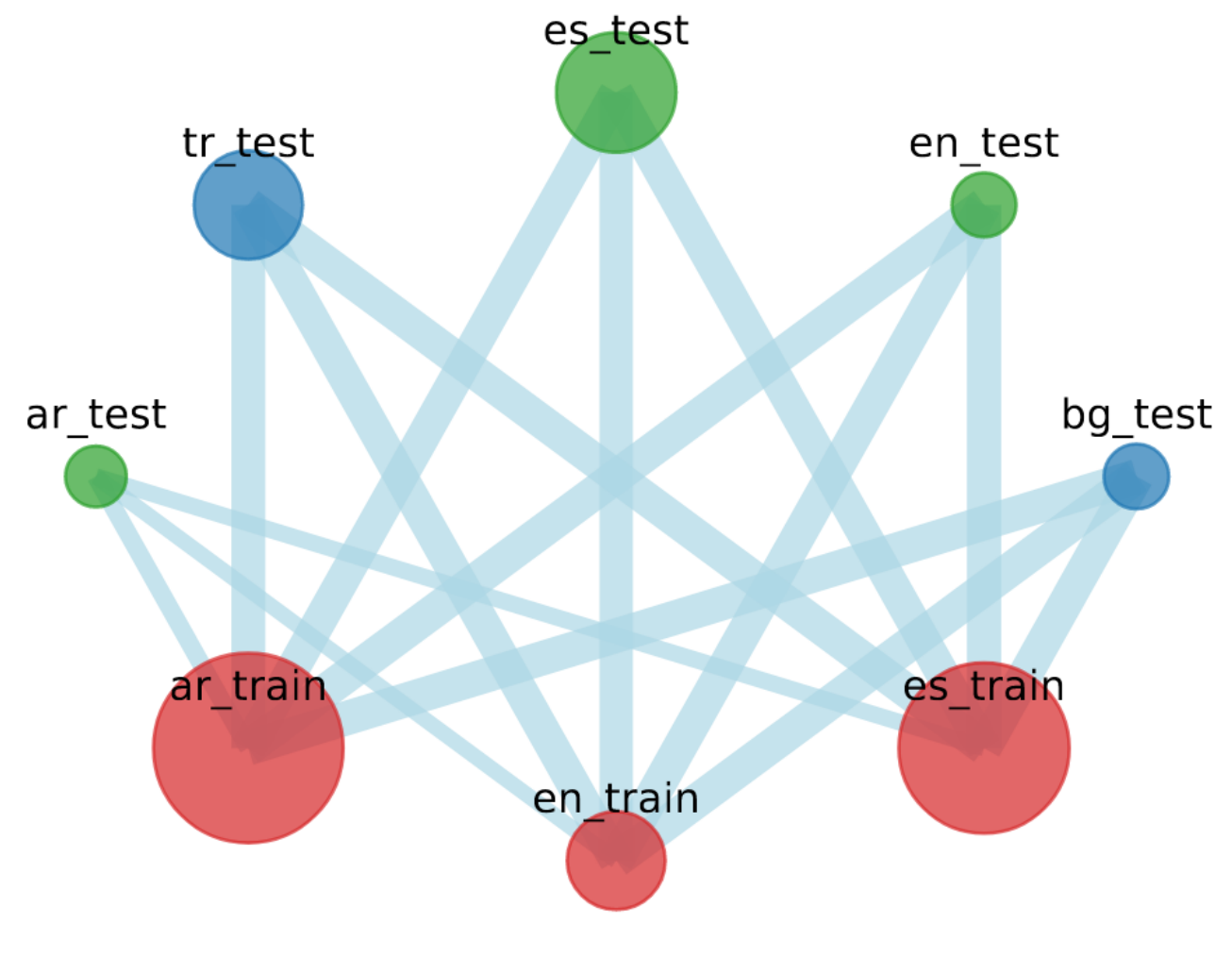}}
    \subfigure[t][mBERT]{\includegraphics[width=0.3\textwidth]{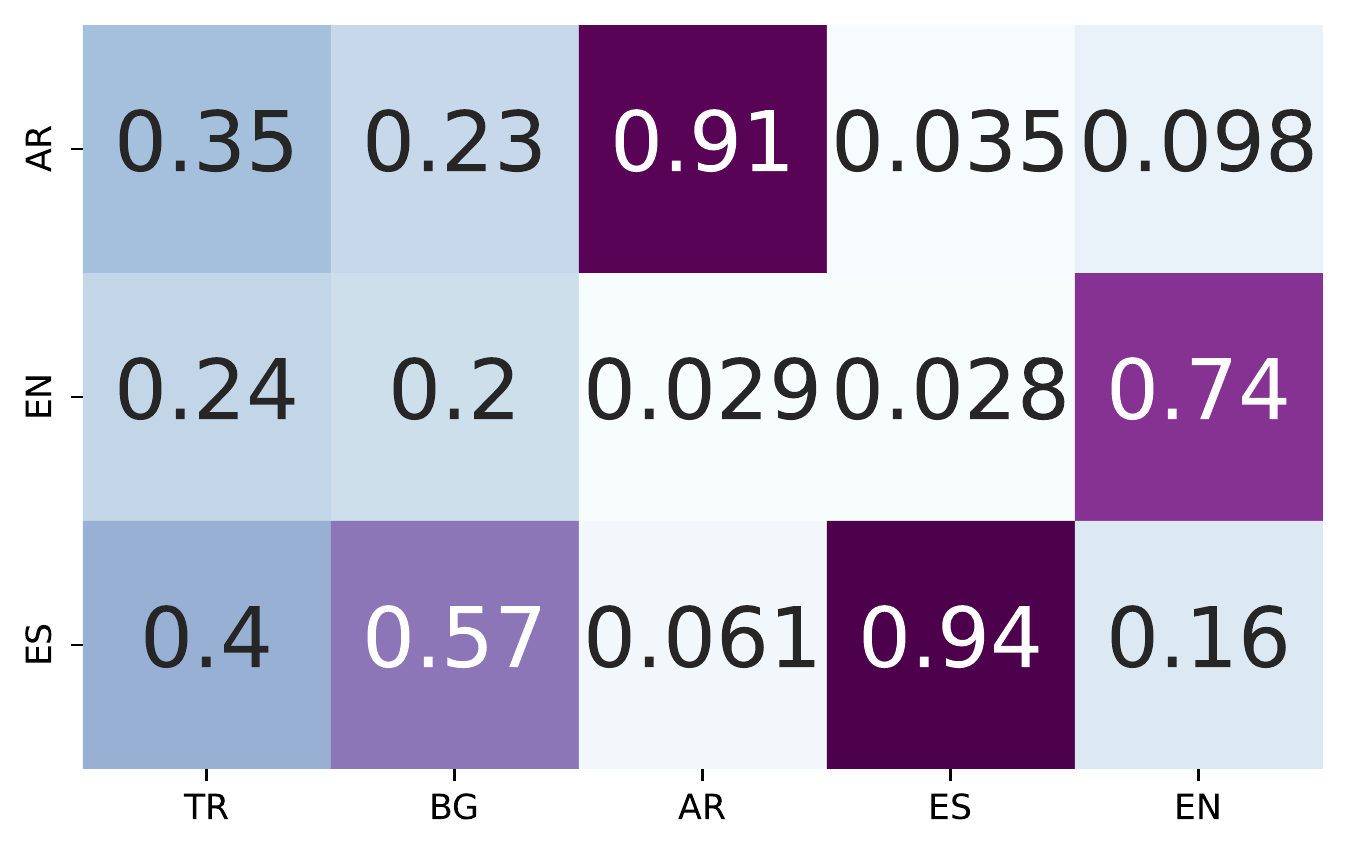}} 
    \subfigure[t][XLM-R]{\includegraphics[width=0.3\textwidth]{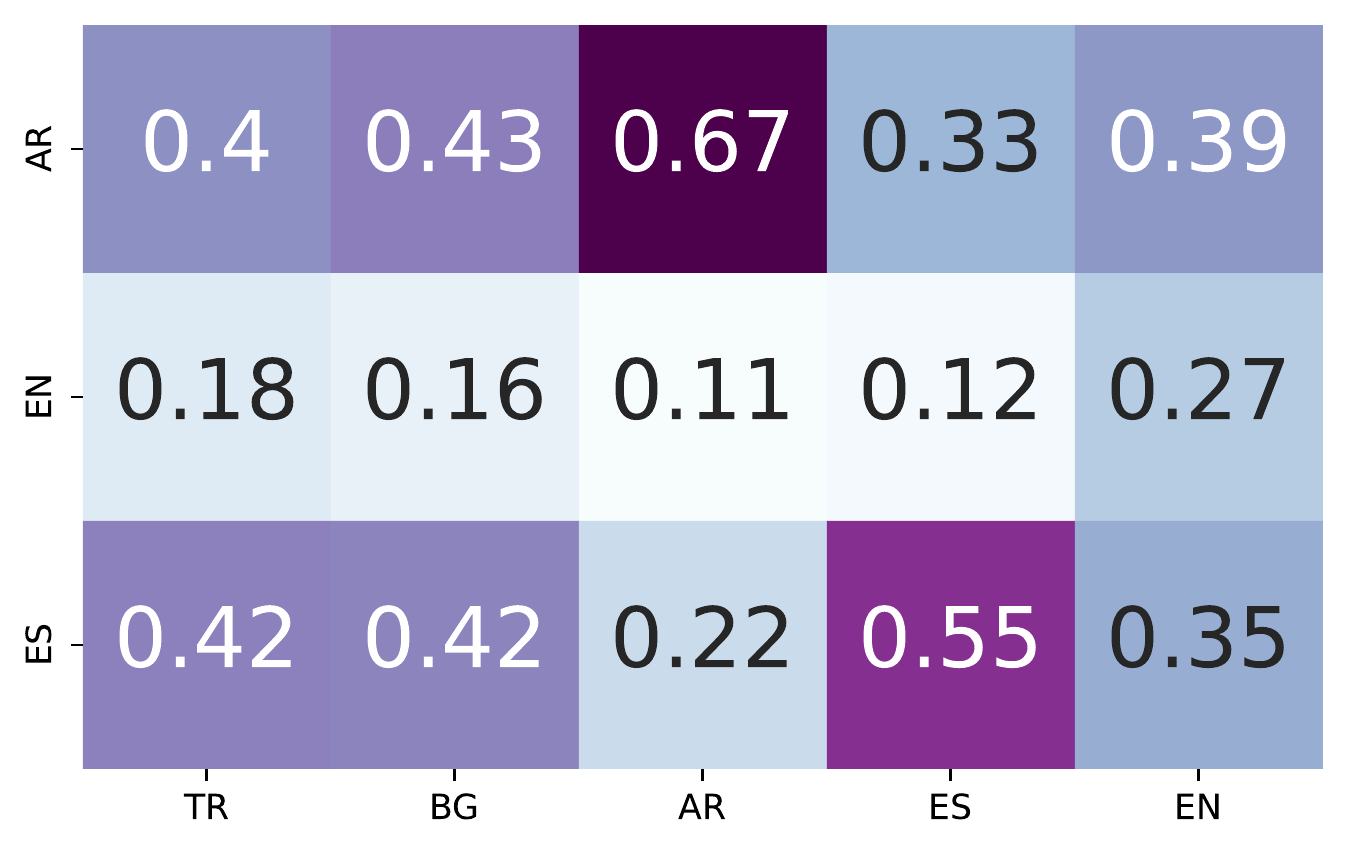}} \hfill
    \subfigure[CT22]{\includegraphics[width=0.25\textwidth] {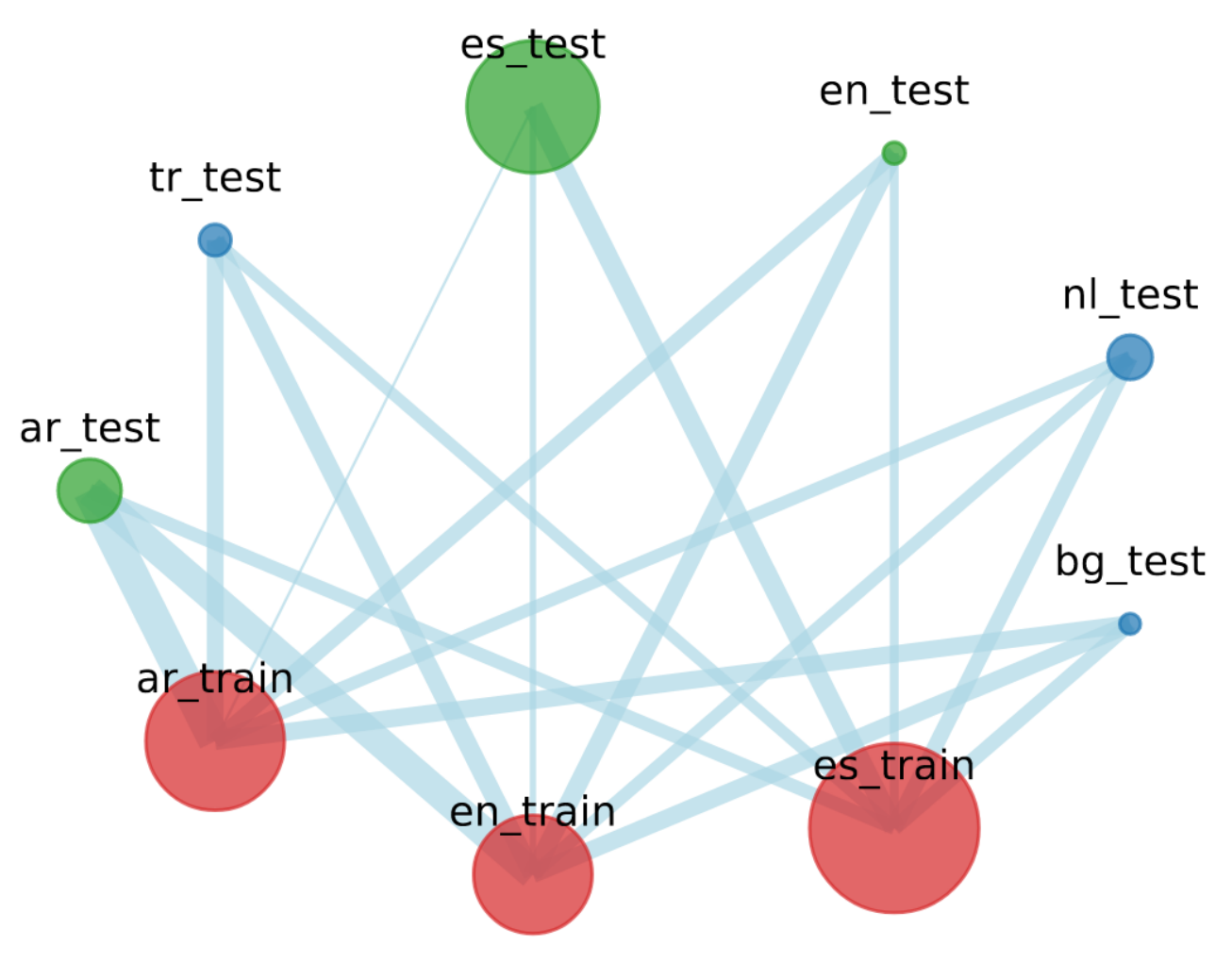}} 
    \subfigure[t][mBERT]{\includegraphics[width=0.3\textwidth]{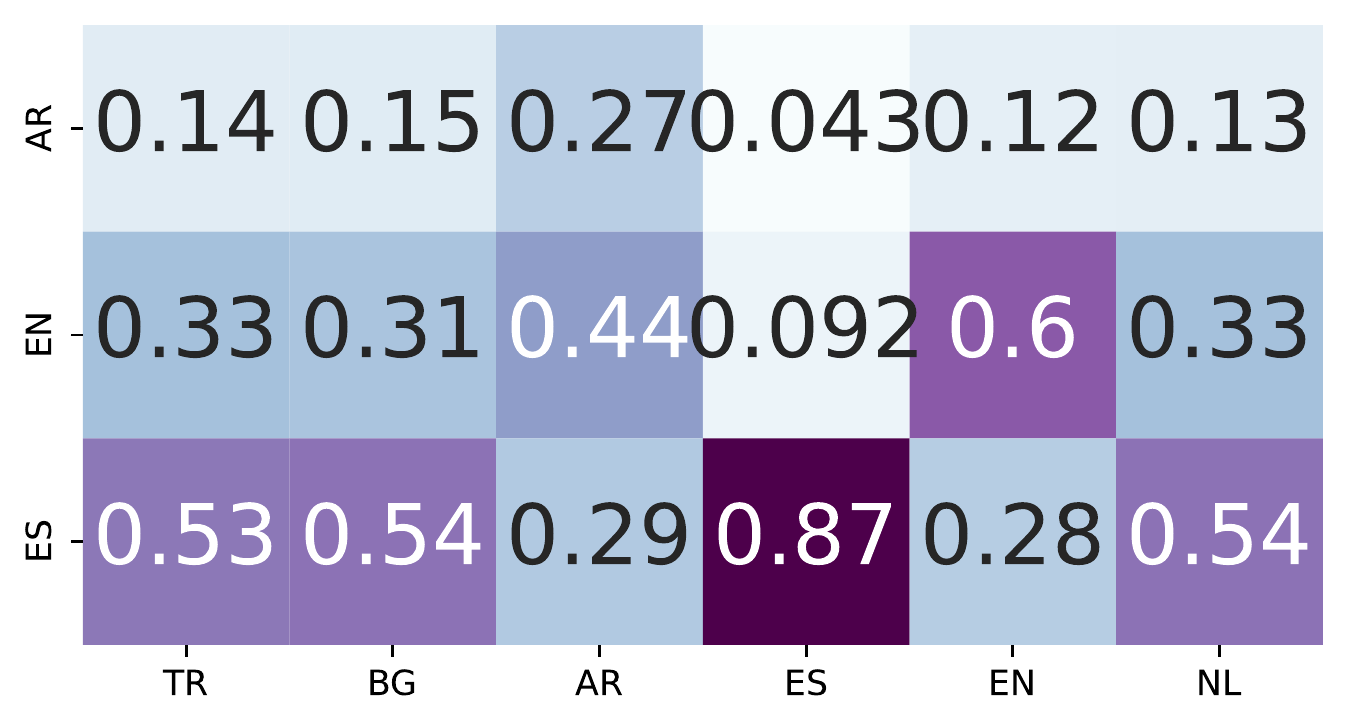}} 
    \subfigure[t][XLM-R]{\includegraphics[width=0.3\textwidth]{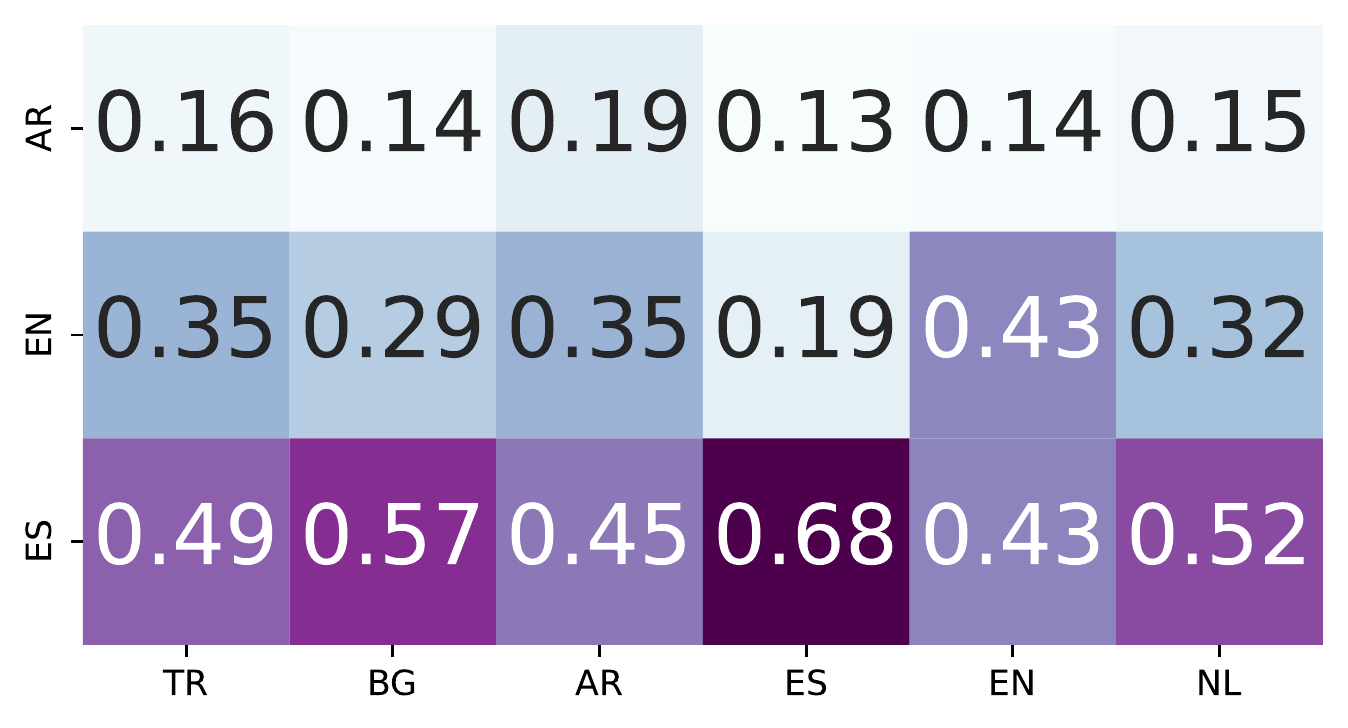}} \\
    \captionsetup{font=small}
    \caption{The left images are topical relation graphs of CT21 (top left) and CT22 (bottom left). \changemarker{In the graphs, the size of the nodes varies by the number of samples, and the edge thickness depends on the overlapping topics}. x-axis of heatmaps shows task adapters, y-axis shows the test samples in the different languages. (b), (c) attention heatmaps of mBERT, (e), (f) attention heatmaps of XLM-R. Topical distribution, and the sample sizes of the \changemarker{training} datasets impact the task adapters' activations. Especially XLM-R TAs are more sensitive than mBERT. }
    \label{fig:heatmaps}
\end{figure*}

\begin{table*}[!t]
    \centering
    \captionsetup{font=small}
    \begin{minipage}[t]{\textwidth}
    \caption{\changemarker{The CW claims predicted correctly by the WL+TA+AF, the examples are in Spanish and Bulgarian. The order of the texts for each example: 1) Visualizations on mBERT, 2) Visualizations on XLM-R 3) the \textcolor{red}{red} text is the translation. The models focus more on GPE (e.g. country names) than the other entity types. We colorized the claims based on their integrated gradients~\cite{DBLP:conf/icml/SundararajanTY17}.}}
    \includegraphics[width=\textwidth]{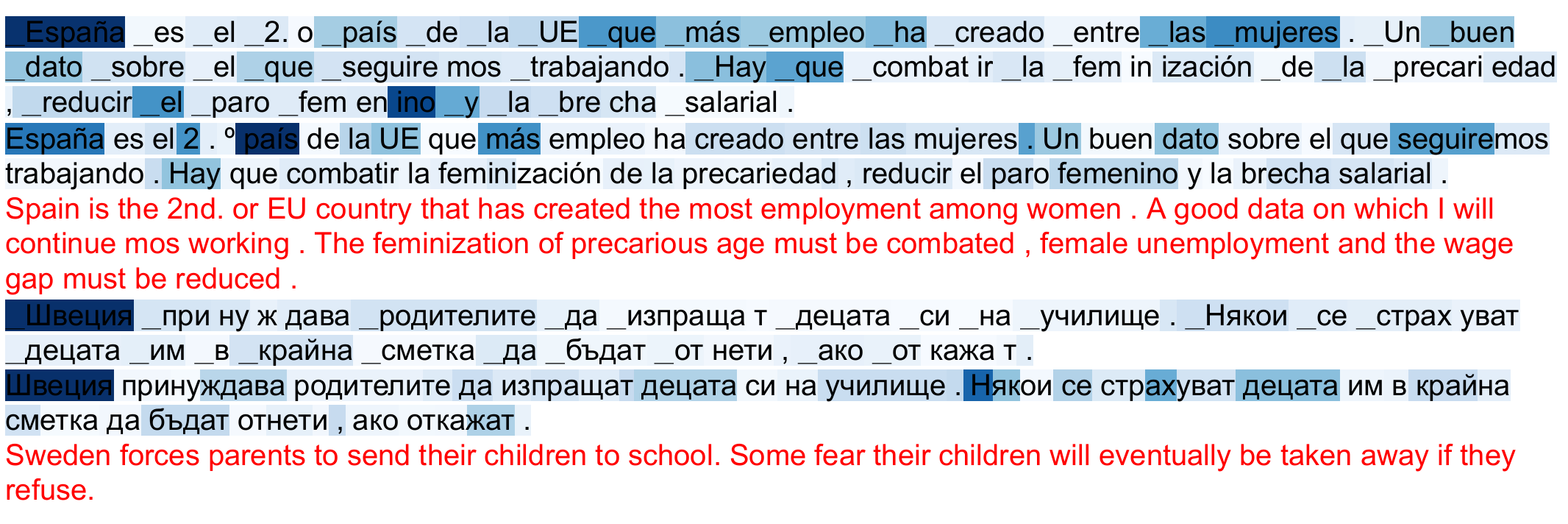}
    \label{fusion_gpe}
    \end{minipage} 
    \begin{minipage}[t]{\textwidth}
    \captionsetup{font=small}
    \caption{\changemarker{The performance of the AF model at predicting entity types in terms of average F1 and the standard deviation. The models could predict GPE more accurately than the others. }}
    \centering
    \scriptsize
    \begin{tabular}{lllll}
    \toprule
    &\textbf{Geo-political Entity} & \textbf{Organization} & \textbf{Number}& \textbf{People} \\
    \textbf{F1} & 46.83 $\pm$ 17.37 & 40.44 $\pm$ 18.62 & 40.99 $\pm$ 22.52 & 37.88 $\pm$ 18.83 \\
    \bottomrule
    \end{tabular}
    \label{tab:results_entities}
    \end{minipage}\vfill
\end{table*}

\section{Further Analysis \label{ablation_study}}

In this section, we present further analysis of the AF models. We investigate AF attentions and then apply an error analysis on the models' predictions.
\\
\noindent
\textbf{Interpretation of the Fusion Attentions.} The AF models can provide an interpretation of which source task adapter might be useful when transferring the knowledge into the target dataset. \changemarker{This kind of analysis would help a data scientist at a newsroom on a decision on which WL should be collected for updating model and managing new resources.} To check the AF behavior on WL+TA+LA+AF, we took the average of the softmax probabilities of the layer of each task adapter in the fusion layer. The higher probability means the more useful the task for determining the label~\cite{DBLP:conf/eacl/PfeifferKRCG21}. In addition, to correlate the attention with the source datasets, we created a graph displaying the topical relationship between source and target sets. In the graph, the nodes are the monolingual datasets; the edges are the overlapped topics between the source and target dataset, weighted by the percentage of the samples about the overlapped topic. The size of nodes are scaled according to sample size.
Figure~\ref{fig:heatmaps} shows the graph for both datasets and the attention weights of mBERT and XLM-R task adapters. Topical distributions and source datasets' size affect which task adapter activates. XLM-R TAs are more sensitive to the source data size and topical relationship. For instance, the Spanish tests in CT22 are weakly connected with the Arabic and English source datasets, and the \changemarker{Spanish TA} of XLM-R has less activation than \changemarker{the mBERT TA}. 

\begin{table*}[!t]
    \captionsetup{font=small}
    \caption{The CW claims that are misclassified by the AF models. The black texts show claims in Turkish. The \textcolor{red}{red} texts are the translations, and the \textcolor{blue}{blue} ones are the explanation of the claims. }
    \centering
    \adjustbox{width=\textwidth}{
    \begin{tabular}{p{16cm}}
    \toprule
         Bunu hep yazdım yine yazacağım , Bakanların aileleri , annesi - babası tam kapsamlı sağlık tedavileri buna ( Estetik dahil ) devlet bütcesinden karşılanıyor da , SMA hastası çocukların tedavisi için niye bir bütçe oluşturulmuyor [UNK] [UNK] \# DevletSMAyıYaşatsın \\
         \textcolor{red}{I've always written this and I will write it again. The families of the ministers, their mother-father full health treatments (including Aesthetics) are covered by the state budget, but why isn't a budget created for the treatment of children with SMA [UNK] [UNK] \# Let The State Live} \\
         \textcolor{blue}{Example of a CW claim that requires local context. SMA is a disease that affects children, and the treatment of SMA is a controversial issue in Turkey.} \\
         Koronavirüs salgınında vaka sayısı 30 bin 021 [UNK] e ulaştı \# Corona \# COVID \# coronavirus \\
         \textcolor{red}{The number of cases in the coronavirus epidemic reached 30 thousand 021 [UNK] \# Corona \# COVID \# coronavirus} \\
         \textcolor{blue}{An example of a CW claim whose veracity could be changed by time.
         }\\
         
    \bottomrule
    \end{tabular}}

    \label{tab:misclassified_samples}
\end{table*}

\noindent
\textbf{Error analysis.} Last, we analyze the misclassified/correctly classified samples by both AF models. \changemarker{As shown in Table~\ref{fusion_gpe}, we noted that the AF model focuses on geo-political entities (GPE). The models could categorize the claims with GPE better than claims containing other type of entities as shown in Table~\ref{tab:results_entities}.}
The importance of the GPE could be learned from the WL corpus whose CW samples have \changemarker{no} negligible amount of these entities (e.g. \%76 of CT21 and \%77 of CT22 Arabic source datasets are GPE). However, the models cannot predict the claims requiring local context, especially in the zero-shot languages. Moreover, the models cannot identify the claims whose veracity can be changed to not CW by time. Some examples are shown in Table~\ref{tab:misclassified_samples}.
\\
\noindent
\textbf{Training efficiency.} \changemarker{We measured the models' training time for one epoch on the datasets. The TA training is on average 4 minutes less than the fully fine tuning. However, the AF training without LAs lasts 3 minutes more, and the training with LAs 9 minutes more than the training time of WL+FFT which was approx. 22 minutes. The methods such as AdapterDrop~\cite{ruckle-etal-2021-adapterdrop} could speed up the AF training.}

\section{Conclusion and Future Work}
In this paper, we investigated the cost efficient cross-training of adapter fusion models on world languages to detect check-worthiness in multiple languages.
The proposed solution performs well on multiple languages, even on zero-shot learning.
Thanks to adapter fusion, the effectiveness of the adapters on particular languages was possible.

The attention of some task adapters seems to depend on the topic and sample distribution in the source dataset.
Ensuring a topical balance across world languages appears to be important.
Our error analysis results indicate that local context is required to detect local claims.
We recommend the usage of background knowledge injection to detect local claims.

In the future, we would like to investigate the injection of background knowledge in adapters and verify our results in additional domains (e.g. war), employing more languages \changemarker{such as German} and focusing on zero-shot learning.

\subsubsection*{Acknowledgements}
We would like to thank the anonymous reviewers, Joan Plepi, Flora Sakketou, Akbar Karimi and Nico Para for their constructive feedback. The work of Ipek Schlicht was part of the KID2 project (led by DW Innovation and co-funded by BKM). The work of Lucie Flek was part of the BMBF projects DeFaktS and DynSoDA. The work of Paolo Rosso was carried out in the framework of IBERIFIER (INEA/CEF/ICT/A202072381931 n.2020-EU-IA-0252), XAI Disinfodemics (PLEC2021-007681) and MARTINI (PCI2022-134990-2). 
\bibliographystyle{splncs04}
\bibliography{custom.bib}
\end{document}